\pdfoutput=1
\documentclass[11pt]{article}

\usepackage[preprint]{acl}

\usepackage{times}
\usepackage{latexsym}
\usepackage[T1]{fontenc}
\usepackage[utf8]{inputenc}
\usepackage{microtype}
\usepackage{inconsolata}
\usepackage{graphicx}
\usepackage{hyperref}
\usepackage{float}
\usepackage{algorithm}
\usepackage{algpseudocode}
\usepackage{amsmath}

\title{SHROOM-INDElab at SemEval-2024 Task 6: Zero- and Few-Shot LLM-Based Classification for Hallucination Detection}

\author{Bradley P. Allen, Fina Polat \and Paul Groth \\
  University of Amsterdam \\
  Amsterdam, NL \\ 
  \texttt{\{b.p.allen, f.yilmazpolat, p.t.groth\}@uva.nl}}

\begin{document}
\maketitle
\begin{abstract}
We describe the University of Amsterdam Intelligent Data Engineering Lab team's entry for the SemEval-2024 Task 6 competition. The SHROOM-INDElab system builds on previous work on using prompt programming and in-context learning with large language models (LLMs) to build classifiers for hallucination detection, and extends that work through the incorporation of context-specific definition of task, role, and target concept, and automated generation of examples for use in a few-shot prompting approach. The resulting system achieved fourth-best and sixth-best performance in the model-agnostic track and model-aware tracks for Task 6, respectively, and evaluation using the validation sets showed that the system's classification decisions were consistent with those of the crowd-sourced human labellers. We further found that a zero-shot approach provided better accuracy than a few-shot approach using automatically generated examples. Code for the system described in this paper is available on Github\footnote{\url{https://www.github.com/bradleypallen/shroom/}}.
\end{abstract}

\section{Introduction}
Prompt engineering of large language models (LLMs) \cite{promptSurvey2023} has recently emerged as a viable approach to the automation of a wide range of natural language processing tasks. Recent work \cite{allen2023conceptual} has focused on the development of zero-shot chain-of-thought \cite{wei2022chain,kojima2022large} classifiers, where hallucination in generated rationales is a concern. Hallucination detection \cite{ji2023survey,huang2023survey} is a way to determine whether the outputs of such systems are sensible, factually correct and faithful to the provided input. The SemEval-2024 Task 6 \cite{mickus-etal-2024-semeval} allows us to evaluate whether and how applying techniques we have developed in the above mentioned work and with related work on knowledge extraction \cite{polat2024testing} using zero- and few-shot classification can provide a means of addressing this concern. Previous systems that perform prompt engineering of LLMs as a means to implement hallucination detection include  SelfCheckGPT \cite{manakul2023selfcheckgpt} and ChainPoll \cite{friel2023chainpoll}.

\begin{figure*}[htbp]
    \includegraphics[width=\textwidth, trim={0 170 0 0}, clip]{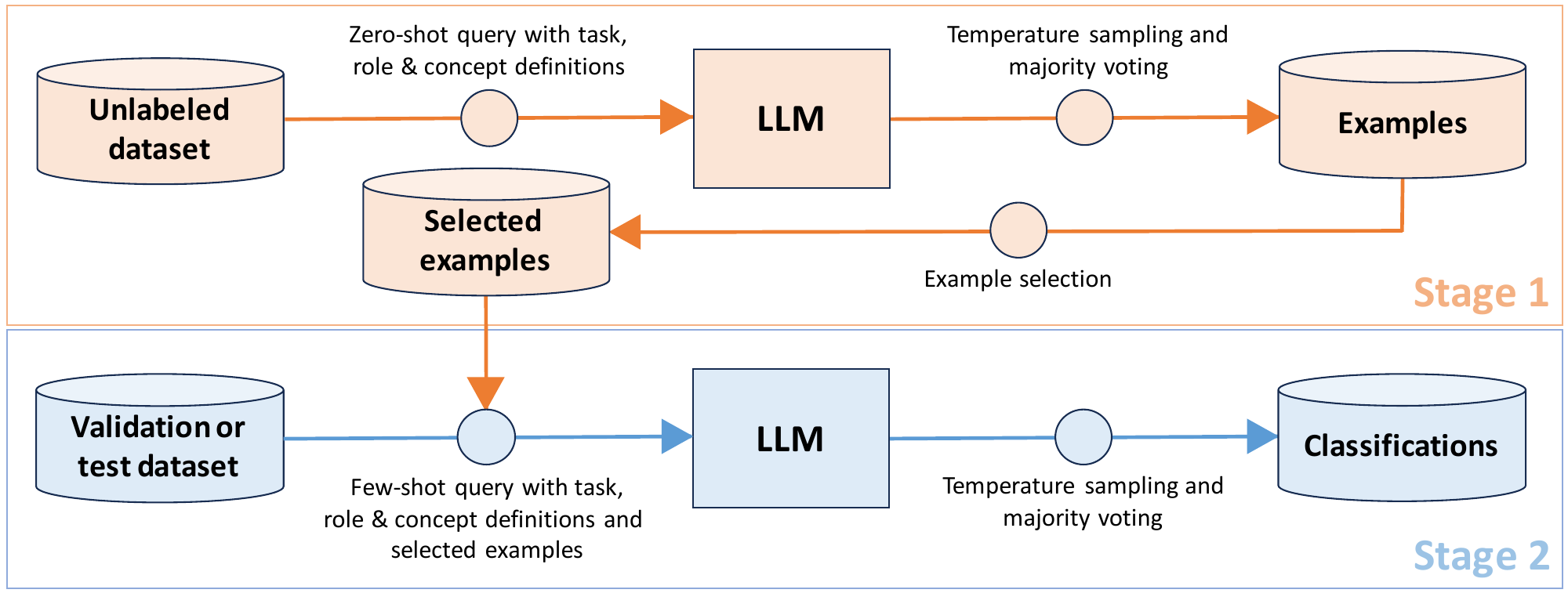}
    \caption{SHROOM-INDElab system workflow.}
    \label{fig:workflow}
\end{figure*}

\section{Data and Task}
The challenge provides a dataset consisting of data points containing: the specific task that a given language model is to perform; an input given to the language model on which to perform that task; a target that is an example of an acceptable output, and the output produced by the language model. Table \ref{tab:datapoint} shows an example of such a data point.

\begin{table}[htbp]
\centering
\resizebox{\columnwidth}{!}{
\begin{tabular}{p{2.5cm}p{8cm}}
\hline
Task &  Definition Modeling\\
Input text &  "The Dutch would sometimes <define> inundate </define> the land to hinder the Spanish army ." \\
Target text &  "To cover with large amounts of water; to flood."\\
Generated text &  "(transitive) To fill with water." \\
\hline
\end{tabular}
}
\caption{Example data point from the unlabeled training dataset for the model-agnostic task.}
\label{tab:datapoint}
\end{table}

Hallucination detection is framed as a binary classification task, where the classifier assigns either 'Hallucination' or 'Not Hallucination' labels with associated probability estimates to data points. Classifier performance is evaluated by comparing these assignments and probabilities to human judgments and their probability estimates, using accuracy and Spearman's correlation coefficient ($\rho$) for assessment. Around 200 crowd-sourced human labellers each labeled about 20 data points. The competition features two tracks: model-agnostic, which uses the basic setup, and model-aware, adding a field for the Hugging Face model identifier of the model generating the text for each data point. Each track provides an unlabeled training dataset and labeled validation and test datasets.

\section{Approach}
Our submission for the SHROOM task is a system that defines classifiers for hallucination detection using prompt engineering of an LLM. Figure \ref{fig:workflow} shows the two-stage workflow used to produce the classifier and evaluate it using the SHROOM datasets.

In Stage 1, we use in-context learning where we ask the LLM to perform the classification according to provided task, role, and concept definition in a zero-shot manner without providing any examples. These classified data points provide examples for a few-shot classifier used in Stage 2. We now proceed to describe the query design and processing steps in the workflow.

\subsection{Zero- and few-shot query design}
\begin{figure*}[htbp]
    \includegraphics[width=\textwidth, trim={60 70 60 0}, clip]{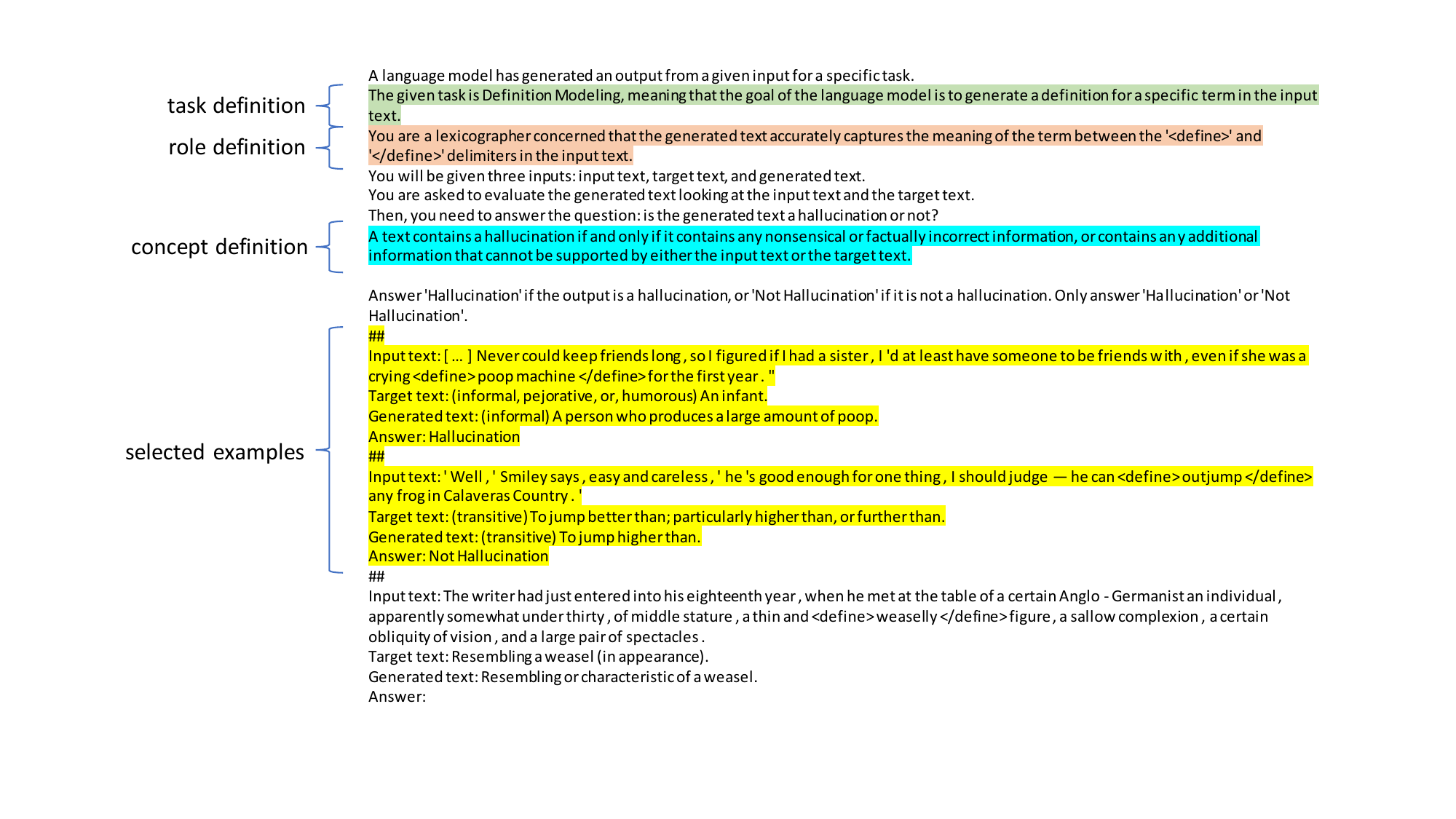}
    \caption{Example prompt for a Stage 2 classifier, given a Definition Modeling task data point from one of the SHROOM datasets, and using 1 example per label.}
    \label{fig:prompt}
\end{figure*}

Figure \ref{fig:prompt} provides an example of the query used to prompt an LLM to produce a classification. The basic prompt template consists of instructions on how to evaluate the generated text according to a hallucination concept definition to answer the question if the generated text is a hallucination or not. Specific guidance is provided such that the form of the answer is in the labels needed to compare directly to the label test data.

The task associated with the data point determines the context for generating both the zero shot and the few shot query based on the prompt template, as illustrated in Figure \ref{fig:prompt}. For the zero-shot query, no examples are included.

The elements involved in instantiating the template given a data point include the task definition performed by another LLM to produce the generated text, a role definition that we assign the classifier to perform, and the concept definition that frames hallucination phenomena and criteria to consider an output as hallucination. The use of role play with LLMs is described by \cite{shanahan2023role} and its use in the context of zero-shot reasoning is described in \cite{kong2023better}. The role definition describes a persona that the LLM is instructed to assume in the context of making a classification decision. For example, for the Definition Modeling task, we instruct the LLM to assume the persona of a lexicographer. The task and role definitions for each task are shown in Table \ref{tab:definitions}. We also provide a single concept definition for the notion of hallucination that is held constant across all of the tasks. 

\begin{table}[htbp]
\centering
\resizebox{\columnwidth}{!}{
\begin{tabular}{p{2.2cm}p{6cm}p{6cm}}
\hline
\textbf{Task} & \textbf{Task definition} & \textbf{Role definition} \\
\hline
Definition Modeling (DM) & The given task is Definition Modeling, meaning that the goal of the language model is to generate a definition for a specific term in the input text. & You are a lexicographer concerned that the generated text accurately captures the meaning of the term between the '<define>' and '</define>' delimiters in the input text. \\
Paraphrase Generation (PG) & The given task is Paraphrase Generation, meaning that the goal of the language model is to generate a paraphrase of the input text. & You are an author concerned that the generated text is an accurate paraphrase that does not distort the meaning of the input text. \\
Machine Translation (MT) & The given task is Machine Translation, meaning that the goal of the language model is to generate a natural language translation of the input text. & You are a translator concerned that the generated text is a good and accurate translation of the input text. \\
Text Simplification (TS) & The given task is Text Simplification, meaning that the goal of the language model is to generate a simplified version of the input text. & You are an editor concerned that the generated text is short, simple, and has the same meaning as the input text. \\
\hline
\end{tabular}
}
\caption{Task and role definitions used for in-context learning.}
\label{tab:definitions}
\end{table}

\subsection{Temperature sampling and majority voting}
Part of the task involves producing an estimate of the probability that a data point exhibits hallucination. In the SHROOM-INDELab system, the estimated probability is calculated by performing temperature sampling \cite{ackley1985learning}, querying the LLM multiple times to generate a sample of classifications, and then dividing the number of positive classifications (i.e., where the generated label is 'Hallucination') by the total number of classifications in the sample. Temperature sampling is performed in producing both Stage 1 zero-shot and Stage 2 few-shot classifications.

\subsection{Example selection}
In Stage 1, the algorithm processes an unlabeled dataset to generate examples using a zero-shot query. Following the Self-Adaptive Prompting approach described in \cite{wan2023better, wan2023universal}, for each task type we sample 64 data points from the unlabelled dataset, and then use a zero-shot query to obtain a classification with estimated probability of hallucination. This information is combined with the data point to produce an example. We partition the examples per task type into two pools, one with positive examples where the label is 'Hallucination' and the other with negative examples where the label is 'Not Hallucination'.

\begin{algorithm}
\small
\begin{algorithmic}[1] 
\Require $P$: generated examples for given task and label, $K$: number of selections
\Ensure  $S$: selected examples 
\State $S \gets \emptyset$ 
\State $Pool \gets P$ 
\For{$k \gets 0$ \textbf{to} $K-1$} 
    \If{$k == 0$} 
        \State $s_k \gets \underset{p \in Pool}{\arg\max}$ $F_0(p)$ 
    \Else 
        \State $s_k \gets \underset{p \in Pool}{\arg\max}$ $F(p, S)$ 
    \EndIf
    \State $S \gets S \cup \{s_k\}$ 
    \State $Pool \gets Pool \setminus \{s_k\}$ 
\EndFor
\end{algorithmic}
\caption{Select examples given a task and label}
\label{alg:selectexamples}
\end{algorithm}

The process used to select the examples to include in the prompt is shown in Algorithm \ref{alg:selectexamples}. The first example chosen from each pool is the one with the maximum negative entropy of the classification probability, as defined in Equation \ref{eq:F0}:
\begin{equation}
\label{eq:F0}
\small
    F_0(p) = p * \log{p} + (1 - p) * \log{(1 - p)}
\end{equation}

For each remaining selection $i \leq K$, the algorithm selects the example that maximizes a trade-off between the diversity of prompts and the consistency of the majority voting result, as defined in Equation \ref{eq:F}: 
\begin{equation}
\label{eq:F}
\small
    F(p, S) = F_0(p) - \lambda \cdot \underset{s \in S}{\max} \left(1 - \text{sim}(\phi(p), \phi(s))\right)
\end{equation}

$\phi$ is calculated for a given data point by concatenating its data into a string and then using an embedding model to produce a representation vector. This trade-off is quantified by subtracting a weighted maximum cosine similarity of the embeddings from the negative entropy, with the weight $\lambda$ controlling the balance between diversity and consistency. In all of our experiments, in keeping with \cite{wan2023universal}, $\lambda$ is set to 0.2. The selected examples for both labels are then serialized and concatenated. This concatenated string is then used to augment the zero-shot query prompt given the task.

\section{Experimental Setup and Results}

\begin{table*}[t!]
\centering
\resizebox{\textwidth}{!}{
\begin{tabular}{p{1.5cm}p{7cm}cccc}
\hline
\multicolumn{2}{c}{ }
& \multicolumn{ 2 }{ c }{ \textbf{model-agnostic} }
& \multicolumn{ 2 }{ c }{ \textbf{model-aware} } \\
\textbf{Dataset} & \textbf{System} & accuracy & $\rho$ & accuracy & $\rho$ \\
\hline
Validation & Baseline & 0.649 (+0.000) & 0.380 (+0.000) & 0.707 (+0.000) & 0.461 (+0.000) \\
& SHROOM-INDElab (\texttt{gpt-3.5-turbo}) & 0.773 (+0.124) & 0.652 (+0.272) & 0.764 (+0.057) & 0.605 (+0.144) \\
& SHROOM-INDElab (\texttt{gpt-4-0125-preview}) & \textbf{0.814} (+0.165) & \textbf{0.697} (+0.317) & \textbf{0.772} (+0.065) & \textbf{0.635} (+0.174) \\
\hline
Test & Baseline & 0.697 (+0.000) & 0.403 (+0.000) & 0.745 (+0.000) & 0.488 (+0.000) \\
& SHROOM-INDElab (\texttt{gpt-4-0125-preview}) & 0.829 (+0.132) & 0.652 (+0.249) & 0.802 (+0.057) & 0.605 (+0.117) \\
& HaRMoNEE & 0.814 (+0.117) & 0.626 (+0.223) & \textbf{0.813} (+0.068) & 0.699 (+0.210) \\
& GroupCheckGPT & \textbf{0.847} (+0.150) & \textbf{0.769} (+0.366) & 0.806 (+0.061) & \textbf{0.715} (+0.227) \\
\hline
\end{tabular}
}
\caption{Classifier performance on SHROOM datasets. $\rho$ = Spearman's correlation coefficient.}
\label{tab:performance}
\end{table*}

The LLMs used in the evaluating the system were from OpenAI (\texttt{gpt-3.5-turbo}, \texttt{gpt-4-0125-preview}) and were invoked using the OpenAI API with the LangChain Python library. Stage 1 was performed once with $K = 5$ using \texttt{gpt-4-0125-preview} on 25 January 2024. The embedding model used in the calculation of $\phi$ was OpenAI \texttt{text-embedding-ada-002}. The Stage 2 run for our final submission during the evaluation period was conducted on 28 January 2024. Runs for the hyperparameter and ablation study results reported below were conducted between 17 February 2024 and 18 February 2024. Approximately \$500 USD in OpenAI API charges were incurred during the above runs.

\subsection{Classification performance}

As shown in Table \ref{tab:performance}, using \texttt{gpt-4-0125-preview} and \texttt{gpt-3.5-turbo} as LLMs our approach showed a significant improvement in both accuracy and Spearman's $\rho$ over the baseline reported for the model-agnostic and model-aware validation sets.\footnote{Although we submitted results for the model-aware track, our implementation of the approach is model agnostic and does not utilize the model field of the data point.}

Our best-performing submission to the competition used \texttt{gpt-4-0125-preview} as its LLM with 1 example provided per label, 20 samples for majority voting, and a temperature setting of 1.2. We compare it to the baseline system's performance on the test datasets together with that reported for each of the first ranked teams in the model-agnostic track (GroupCheckGPT) and the model-aware track (HaRMoNEE). The SHROOM-INDElab system ranked fourth and sixth in the tracks, respectively.

The values of $\rho$ can be interpreted as showing a moderate to strong correlation between the estimated probability of hallucination provided by the system and that provided by the majority vote result of the human labellers.

\subsection{Hyperparameter study}
The classifier has three hyperparameters; temperature, which is the parameter passed to the language model to indicate the level of stochasticity associated with its generation process; the number of examples per label provided for in-context learning; and the number of samples per query performed and used to calculate the estimated probability associated with the classification of the data point.

We investigated the impact of varying the values of the three hyperparameters of the classifier on the classifier's performance. We used \texttt{gpt-3.5-turbo} to conduct this investigation, computing values of accuracy and Spearman's $\rho$ by executing three different passes over the model-agnostic validation dataset.

\begin{figure}[ht!]
    \includegraphics[width=\columnwidth, trim={0 0 0 0}, clip]{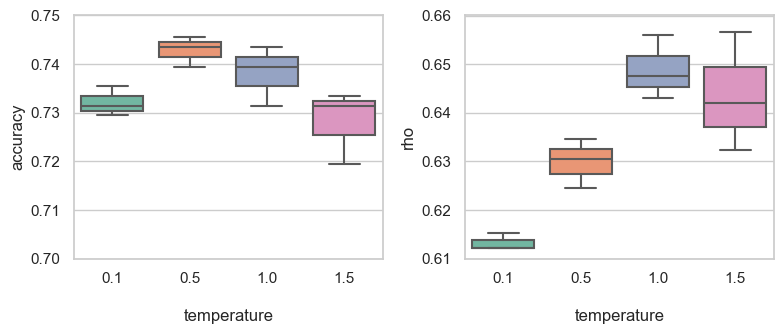}
    \caption{Classifier performance by temperature.}
    \label{fig:temperature}
\end{figure}

Figure \ref{fig:temperature} shows the best classifier accuracy is obtained with a temperature between 0.5 and 1.0, and that the best value for Spearman's $\rho$ is obtained with a temperature between 0.5 and 1.5, given settings of 1 example per label and 5 samples per query.

\begin{figure}[ht!]
    \includegraphics[width=\columnwidth, trim={0 0 0 0}, clip]{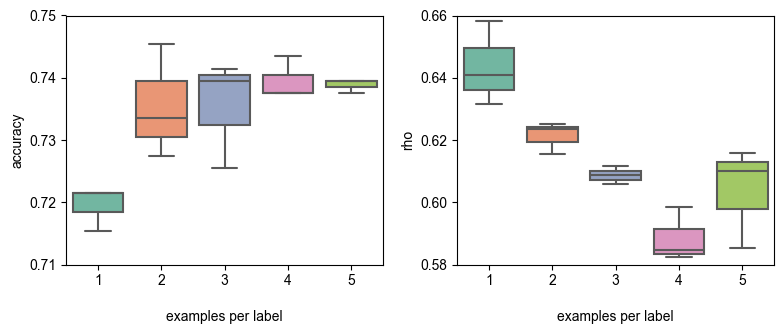}
    \caption{Classifier performance by examples per label.}
    \label{fig:examples-per-label}
\end{figure}

Figure \ref{fig:examples-per-label} shows that increasing the number of examples for few-shot classification beyond one per label led to an increase in accuracy with diminishing returns after 2 examples per label, but a decrease in Spearman's $\rho$, given settings of temperature of 1.0 and 5 samples per query.

\begin{figure}[ht!]
    \includegraphics[width=\columnwidth, trim={0 0 0 0}, clip]{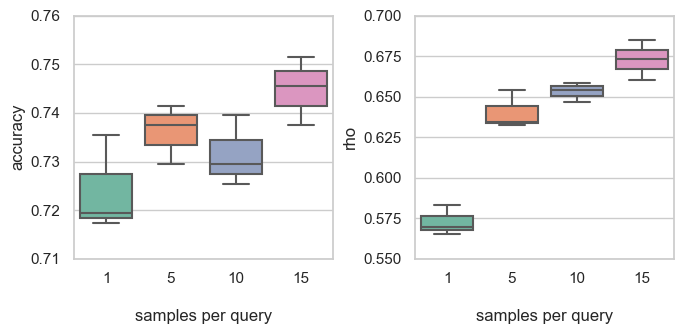}
    \caption{Classifier performance by samples per query.}
    \label{fig:samples-per-query}
\end{figure}

Figure \ref{fig:samples-per-query} shows that increasing the number of samples per query led to an increase in both accuracy and Spearman's $\rho$, given 1 example per label and a temperature of 1.0.

\subsection{Ablation study}
\begin{figure}[htbp]
    \includegraphics[width=\columnwidth, trim={0 0 0 0}, clip]{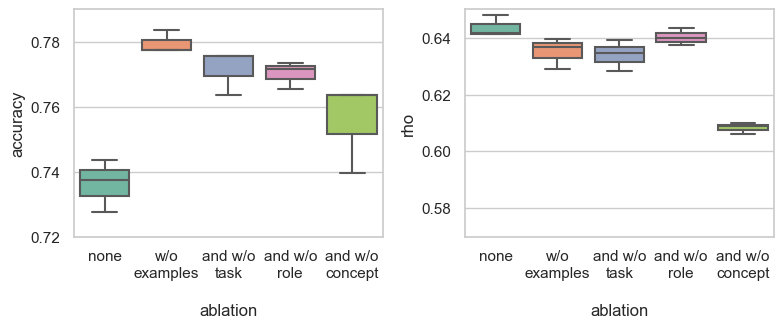}
    \caption{Ablation study using the model-agnostic validation dataset.}
    \label{fig:ablations}
\end{figure}

Figure \ref{fig:ablations} shows the results of an ablation study to determine the contribution of the various elements of the prompt provided to the language models. We evaluated the contribution of each of the components of the Stage 2 classifier prompt by removing each in sequence, in the following order: the selected examples, the task definition, the role definition, and finally the concept definition. The ablation study was conducted using \texttt{gpt-3.5-turbo}, with 1 example per label, 5 samples per query, and a temperature of 1.0, again involving three different passes over the model-agnostic validation dataset.

We interpret the results of the ablation study as indicating that the use of examples led to poorer accuracy but slightly better Spearman's $\rho$, that the contributions of the definitions of task and role towards classifier performance were minimal, but that the contribution of the definition of the concept of hallucination was significant.

\subsection{Level of agreement with human labellers}

We also investigated the degree of inter-annotator alignment exhibited with respect to the model-agnostic data set. Based on the human labeling data associated with each data point in the model-agnostic validation data set, we obtained a Fleiss' $\kappa$ of 0.373, which can be interpreted as indicating a fair level of agreement among the human labellers, which in turn implies that the reliability of the human labeling might be reasonable, but is not highly consistent or unanimous. Adding the classifier's labeling yields an increase in Fleiss' $\kappa$ to 0.405, closer to a moderate level of agreement, which implies that the classifier's decisions are consistent with those of the human labellers. 

\begin{table}[htbp]
\centering
\resizebox{\columnwidth}{!}{
\begin{tabular}{p{3.5cm}cccc}
\hline
\textbf{human consensus} & \textbf{N} & \textbf{accuracy} & \textbf{$\kappa$} & \textbf{$\rho$} \\
\hline
low (2/3 split) & 145 & 0.621 & 0.238 & 0.224 \\
high (4/5 split) & 171 & 0.854 &0.701 & 0.734 \\
unanimous & 183 & \textbf{0.929} & \textbf{0.856} & \textbf{0.885} \\
\hline
all & 499 & 0.814 & 0.623 & 0.697 \\
\hline
\end{tabular}
}
\caption{Alignment between the system and human labellers.}
\label{tab:iiaresults}
\end{table}

We then proceeded to investigate the relationship between the degree of agreement between human labellers and system performance. Table \ref{tab:iiaresults} shows the level of agreement between the system and the human labellers, as measured by taking subsets of data points from the model-agnostic validation dataset filtered by the three degrees of consistency in human labeling and calculating the pairwise Cohen's $\kappa$ between the system's labeling and the label provided by taking the majority vote of the human labellers. The results indicate that system agreement with human labeling increases as the certainty of the human labeling increases.

\section{Discussion and Conclusion}
In summary, the SHROOM-INDElab system was competitive with the other systems submitted for evaluation, and system labeling was consistent with that of human labellers. 

The result in the ablation study that the exclusion of selected examples led to better accuracy suggests the need for further investigation with respect to how the way in which examples are selected and included in the classifier prompts impacts accuracy to determine the cause of the problem. The result that the exclusion of an explicit definition of hallucination leads to poorer accuracy and Spearman's $\rho$ suggests the utility of including intentional definitions of concepts in prompts for LLM-based classifiers \cite{allen2023conceptual}.  

Given the above results, we plan to investigate the use of this approach to hallucination detection in future work on the evaluation of natural language rationale generation \cite{li2024leveraging} in the context of zero- and few-shot chain-of-thought classifiers for use in knowledge graph evaluation and refinement \cite{allen2023knowledge}.

\section*{Acknowledgements}
This work is partially supported by the European Union’s Horizon Europe research and innovation programme within the ENEXA project (grant Agreement no. 101070305).

\bibliography{main} 

\end{document}